\documentclass{article}
\usepackage{spconf}
\usepackage{hyperref}
\usepackage{graphicx}
\usepackage{subfigure}
\usepackage{amsmath,amsfonts,amssymb}
\usepackage{cleveref}
\usepackage{booktabs}
\usepackage{multirow}
\usepackage[numbers,sort&compress]{natbib}

\usepackage{enumitem}
\setlist{nosep, leftmargin=14pt}

\usepackage{amsmath,amsfonts,bm}

\def\eqref#1{equation~\ref{#1}}

\def\1{\bm{1}}

\def\vg{{\bm{g}}}

\def\vw{{\bm{w}}}
\def\vx{{\bm{x}}}

\DeclareMathAlphabet{\mathsfit}{\encodingdefault}{\sfdefault}{m}{sl}
\SetMathAlphabet{\mathsfit}{bold}{\encodingdefault}{\sfdefault}{bx}{n}

\def\gD{{\mathcal{D}}}

\def\gX{{\mathcal{X}}}
\def\gY{{\mathcal{Y}}}

\def\sR{{\mathbb{R}}}

\newcommand{\E}{\mathbb{E}}

\title{Learning the irreversible progression trajectory of Alzheimer's disease}

\name{Yipei Wang$^\dagger$, Bing He$^\ddagger$, Shannon Risacher$^\ddagger$, Andrew Saykin$^\ddagger$, Jingwen Yan$^{\ddagger*}$, Xiaoqian Wang$^{\dagger*}$\thanks{* Co-corresponding authors. J. Yan and X. Wang were supported by NSF award 1955890, 2146091, 1942394, and NIH R21AG072101, U19AG074879, R01LM013463 and U01AG068057.}
}
\address{$^\dagger$Purdue University, $^\ddagger$Indiana University}

\begin{document}

\onecolumn
\noindent
© 20XX IEEE.  Personal use of this material is permitted.  Permission from IEEE must be obtained for all other uses, in any current or future media, including reprinting/republishing this material for advertising or promotional purposes, creating new collective works, for resale or redistribution to servers or lists, or reuse of any copyrighted component of this work in other works.

\twocolumn
\maketitle
\begin{abstract}
Alzheimer's disease (AD) is a progressive and irreversible brain disorder that unfolds over the course of 30 years. Therefore, it is critical to capture the disease progression in an early stage such that intervention can be applied before the onset of symptoms. Machine learning (ML) models have been shown effective in predicting the onset of AD. 
Yet for subjects with follow-up visits, existing techniques for AD classification only aim for accurate group assignment, where the monotonically increasing risk across follow-up visits is usually ignored. Resulted fluctuating risk scores across visits violate the irreversibility of AD, hampering the trustworthiness of models and also providing little value to understanding the disease progression. To address this issue, we propose a novel regularization approach to predict AD longitudinally. Our technique aims to maintain the expected monotonicity of increasing disease risk during progression while preserving expressiveness. Specifically, we introduce a monotonicity constraint that encourages the model to predict disease risk in a consistent and ordered manner across follow-up visits. 
We evaluate our method using the longitudinal structural MRI and amyloid-PET imaging data from the Alzheimer's Disease Neuroimaging Initiative (ADNI). Our model outperforms existing techniques in capturing the progressiveness of disease risk, and at the same time preserves prediction accuracy.
\end{abstract}


\vspace{-5pt}
\section{Introduction}
\vspace{-5pt}
\label{sec:intro}

Alzheimer's disease (AD) is a complex irreversible neurodegenerative disease that affects cognitive functions, including memory, thinking, and behaviors. It is the most common form of dementia, accounting for more than half of the cases. AD symptoms evolve progressively with age and may take up to 30 years to unfold \citep{hampel2021amyloid}. Efforts are increasing for early diagnosis of AD to enable timely intervention before symptom onset, aiming to stop or slow down disease progression.

Current staging of AD includes healthy control (HC), early mild cognitive impairment (EMCI), late MCI (LMCI), and AD. Due to the extremely long spectrum of AD, the available longitudinal imaging data offers only a snapshot of each patient at one or a few visits. 

\begin{figure}[t]
    \centering
    \includegraphics[width=\columnwidth]{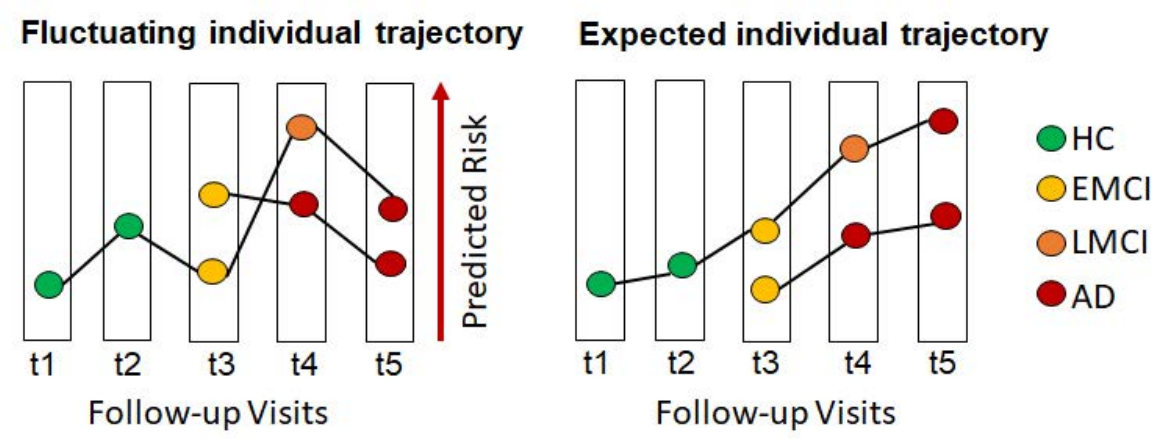}
    \caption{Illustration of fluctuating and expected individual progression trajectories across follow-up visits. Each connected line indicates one subject with multiple visits. Left: fluctuating trajectories commonly seen in raw data and from existing classification models which rely on baseline data only. Right: Expected individual trajectory with predicted risk monotonically increasing across follow-up visits.}
    \vspace{-5pt}
    \label{fig:intro_trajectory}
\end{figure}

Existing classification studies have achieved great success in differentiating HC from AD, yet a significant gap of progression between the two stages remains to be learned \citep{tong2017multi,liu2012ensemble,wisnu2022xadlime, wang2022deep}. There has been a recent shift of effort to differentiate HC from EMCI or differentiate stable MCIs and MCI converters, with very limited success so far though \citep{basaia2019automated}. While existing work mainly focuses on group progression \citep{ouyang2020longitudinal,wang2022deep,tasaki2022inferring}, trajectories formed by connecting predictions of same subject will likely show capricious trends over time (\cref{fig:intro_trajectory} Left), since visits of the same subjects are taken as independent inputs. This contradicts the irreversible progression of AD and thereby harms the trustworthiness of models in real-world applications.

Prior knowledge has been demonstrated very effective in regularizing the models for desired behaviors \citep{ross2017right,rieger2020interpretations,boopathy2020proper,wang2021self}. Therefore, with the aim of enhancing the trustworthiness of models in predicting AD stages, we leverage the irreversibility of AD as prior and propose a new regularization approach to help model individual progressiveness across follow-up visits. The proposed framework is evaluated using the longitudinal imaging data from the Alzheimer's disease Neuroimaging Initiative (ANDI) datasets including structural magnetic resonance imaging (MRI) and positron emission tomography (PET) for amyloid deposition. Our experiments demonstrated that when this prior knowledge is imposed, the trade-off to the expressiveness of the model is negligible, but the gain in the desired behaviors (i.e., generating expected individual progression trajectory like (\cref{fig:intro_trajectory} Right)) is invaluable.

\vspace{-10pt}
\section{Methods}
\vspace{-10pt}
\label{sec:methods}
\subsection{Datasets} 
\vspace{-5pt}
We downloaded longitudinal structural MRI, amyloid PET and other clinical data from the Alzheimer’s Disease Neuroimaging Initiative (ADNI) database (https://adni.loni.usc.edu/). The ADNI is a longitudinal study launched in 2003 to track the progression of AD by using clinical and cognitive tests, MRI, FDG-PET, amyloid PET, CSF, and blood biomarkers. More details can be found in previous reports\citep{jack2010update,saykin2010alzheimer}. The study population was composed of participants from the ADNI-1, ADNI-2, and ADNI-GO\citep{weiner2013alzheimer}. Subjects with reversed diagnoses in follow-up visits, such as conversion from AD back to EMCI, were excluded from this study. In total, we have 7702 data points from 1793 subjects for structural MRI, and 2377 data points from 1054 subjects for amyloid PET. For both modalities, summary measures from brain regions of interest (ROI) were directly obtained from the ADNI. For structural MRI, volumes of 16 subcortical ROIs and thickness of 68 cortical ROIs were included. For amyloid PET, standardized uptake value ratio (SUVR) of 68 cortical ROIs, indicating the level of amyloid deposition, were included and further normalized using \texttt{COMPOSITE\_REF\_SUVR} (a summary measure provided by ADNI) as reference. Subcortical regions were excluded for amyloid analysis since their amyloid burden is found non-specific and not related to AD risk \citep{edmonds2016patterns}. Using the weight derived from baseline HC subjects, all imaging measures were pre-adjusted to remove the potential bias introduced by age, gender and years of education. Intracranial volume (ICV) was used as an additional covariate for volume and thickness measures. \cref{tab:tab_demo} shows the detailed demographic information, where age is taken from baseline (BL) (i.e., first visit).

\begin{table}[htbp]
  \centering
  \scriptsize
  \vspace{-10pt}
  \caption{Demographic information of all participants.
  }
    \begin{tabular}{ccccc}
    \toprule
            MRI     &  HC   & EMCI  & LMCI  & AD \\
    \midrule
           Time points & 2284  & 1427  & 2230  & 1761 \\
           Subject number & 466   & 361   & 590   & 361 \\
           Gender(M/F) & 222/244 & 205/156 & 367/223 & 204/157 \\
           Age(mean$\pm$sd) & 73.95$\pm$5.94 & 71.56$\pm$7.32 & 74.07$\pm$7.6 & 75.14$\pm$7.76 \\
           Educ(mean$\pm$sd) & 16.48$\pm$2.59 & 16.01$\pm$2.61 & 15.96$\pm$2.84 & 15.3$\pm$2.89 \\
    \midrule
           Amyloid-PET     &  HC   & EMCI  & LMCI  & AD \\
    \midrule
           Time points & 791   & 738   & 424   & 424 \\
           Subject number & 200   & 316   & 176   & 152 \\
           Gender(M/F) & 86/114 & 180/136 & 100/76 & 91/61 \\
           Age(mean$\pm$sd) & 73.04$\pm$6.26 & 71.73$\pm$7.16 & 72.49$\pm$7.52 & 74.78$\pm$8.21 \\
           Educ(mean$\pm$sd) & 16.58$\pm$2.47 & 16.08$\pm$2.64 & 16.57$\pm$2.5 & 15.67$\pm$2.63 \\
    \bottomrule
    \end{tabular}%
  \label{tab:tab_demo}%
\end{table}%

All imaging features were standardized to zero mean and unit standard deviation for subsequent analysis. Each data point is associated with a label of group HC/EMCI/LMCI/AD. Due to limited AD data and the difficulties in distinguishing MCI stages, the classification tasks of AD stages usually focus on binary subtasks \citep{oh2019classification,zhou2019effective,you2020alzheimer,odusami2021analysis}. We consider the binary problems including the three consecutive stages: (1) HC/EMCI; (2) EMCI/LMCI; (3) LMCI/AD, and (4) HC/AD, which classifies samples of completely healthy and dementia.

\vspace{-10pt}
\subsection{Model Formulation}
\vspace{-5pt}

We consider the dataset $\gD = \{(X_i, Y_i)\}_{i=1}^M$ of $M$ subjects, where $Y_i$ denotes the stages HC/EMCI/LMCI/AD. Each subject contains a series of longitudinal data $X_i = [\vx_{i,1},\cdots,\vx_{i,K_i}]^T$, where $\vx_{i,k}\in\sR^{d}$ is a single data point consisting of $d$ features. Therefore the entire dataset can also be written as $\gD = \gX\times\gY\in\sR^{(\sum_{i=1}^MK_i)\times d}\times\sR^N$.

A classifier for such data is denoted as $f_\theta:\sR^d\rightarrow\sR$, parameterized by $\theta$.
The ERM classifier is obtained by $\theta^* = \arg\min_{\theta} L_{\textrm{cls}}(\theta)$,
where $L_{\textrm{cls}}$ is the classification loss.
For the tabular data of MRI and amyloid-PET, various models are applicable for the prediction of different stages, including non-deep models \citep{shwartz2022tabular} such as linear logistic regressions (LR) and the variants Lasso, Ridge, elastic net, linear discriminant analysis (LDA), random forests, XGBoost \citep{chen2016xgboost}, etc. Recent studies show that deep neural networks such as multilayer perceptron (MLP) \citep{rumelhart1985learning} can also be as powerful as, or even outperform non-deep models on tabular tasks \citep{gorishniy2021revisiting}. In fact, deep models have been extensively used in both supervised and unsupervised tasks for AD \citep{zhou2019effective,ebrahimighahnavieh2020deep,wen2020convolutional,saleem2022deep}.
When $f$ is a deep model, it can be decomposed as $f(\vx) = \vw^T\vg(\vx)+b$, where $\vg:\sR^d\rightarrow\sR^p$ is the feature extraction from the input data to the penultimate output before the last linear classification layer. And $\vw\in\sR^p,b\in\sR$ represent the weights and the bias terms of the last linear layer.

\vspace{-10pt}
\subsection{Irreversible Subject Trajectories}
\vspace{-5pt}

The progression of the disease is known as irreversible and should always follow the continuum of AD for all subjects. Therefore, given subject $i$ with data $X_i = [\vx_{i,1},\cdots,\vx_{i,K_i}]^T$, $Y_i = [y_{i,1},\cdots,y_{i,K_i}]^T$, the predictions of all samples are expected to follow the trajectory. That is, $f(\vx_{i,j})\le f(\vx_{i,j+1})$ for all $\forall j,1<j<{K_i-1}$. Note that the binary labels are indeed monotonic, $y_{i,j}\le y_{i,j+1}$. Nonetheless, since the labels are discrete, achieving high accuracy does not actually contradict the frequent breaks of the monotonicity. Even if with 100\% accuracy such that $\bm{1}_{f(\vx_{i,j}) > 0} = y_{i,j}$ holds for all $\forall j,1<j<K_i$, the trajectory can still be nowhere monotonically increasing except for when the subjects deteriorate into the next stage. In fact, we carry out experiments in \cref{sec:experiments} that show the prediction trajectories can be very capricious without regularization.

\vspace{-10pt}
\subsection{Regularization}
\vspace{-5pt}

\textbf{Neighbor Regularization.} In order to achieve the monotonic prediction such that $f(\vx_{i,k})\le f(\vx_{i,k+1})$, it suffices to require $f(\vx_{i,k+1}) - f(\vx_{i,k}) \ge 0$. 
However, the term is unbounded and cannot be maximized.
Notice that through the last linear classification layer, it is equivalent to $\vw^T\big(\vg(\vx_{i,k+1})-\vg(\vx_{i,k})\big)$. 
We thus consider the cosine similarity for bounded loss term, which results in the following neighbor regularization bounded in $[-K_i,K_i]$. 

\vspace{-10pt}
\begin{align}
    L_{nb}^i = -\sum_{k<K_i}\frac{\vw^T(\vg(\vx_{i,k+1})-\vg(\vx_k))}{\|\vw\|\|\vg(\vx_{i,k+1})-\vg(\vx_k)\|}
\end{align}
\vspace{-10pt}

\noindent\textbf{Complete Regularization.} 
Although $L_{nb}$ implicitly requires overall monotonicity, the penalty for one extreme violation is overlooked. Even if $f(\vx_{i,K_i})<f(\vx_{i,k})$ for $\forall k<K_i$, the penalty only exists in the last term and is at most 1. This leads to very unstable embedding mapping $\vg$. In order to resolve this, a stronger regularization takes 
all sample pairs $(k_1,k_2)$ into consideration by their temporal information and is bounded in $[-\frac{K_i(K_i-1)}{2}, \frac{K_i(K_i-1)}{2}]$. Besides, we expect the more temporally distant the two samples are, the more strictly they should obey such monotonicity. Therefore each pair is weighted by the time span associated, denoted by $\tau_i(k_2) - \tau_i(k_1)$, where $\tau_i(k)\in\sR_+$ denotes the age of the $i$-th subject associated with sample $\vx_{i,k}$. Therefore, the regularization term for all subjects can be written as $\min_{\theta}L_{reg}$ where $L_{reg}$ is the expected regularizer over the entire dataset

\vspace{-10pt}
\begin{align}
    \begin{split}
    L_{reg}\approx -\sum_{i=1}^M\Big[\sum_{k_1<k_2\le K_i}&\frac{\vw^T(g(\vx_{i,k_2})-g(\vx_{i,k_1}))}{\|\vw\|\|g(\vx_{i,k_2})-g(\vx_{i,k_1})\|}\\
    &\big(\tau_i(k_2) - \tau_i(k_1)\big)\Big]/M
    \end{split}
\end{align}
\vspace{-10pt}

In practice, we randomly sample a small batch of subjects for MRI data as the size of the dataset is much larger. For amyloid-PET data, we regularize the entire dataset in each iteration. The loss function is then written as
$L(\theta) = L_{\textrm{cls}}(\theta) + \gamma L_{reg}(\theta)$,
where $\gamma=2\times 10^{-4}$ is the balance coefficient.

\noindent\textbf{Evaluation of Monotonicity.} For the purpose of evaluating the monotonicity of model $f$ w.r.t. the dataset $\gX$, we count the \# of violation pairs for each subject and use the expected ratio of the \# of violations to the \# of all pairs. That is,

\vspace{-10pt}
\begin{align}
    r(f,\gX) =& \E_{X\sim\gX}\Big[\frac{1}{\binom{|X|}{2}}\sum_{k_1<k_2\le|X|}\bm{1}_{[f(X_{k_2})<f(X_{k_1})]}\Big]
\end{align}
\vspace{-10pt}

The ratio $r\in[0,1]$ is expected to be the smaller the better, and it reaches $r=0$ if and only if the $f$ is nowhere decreasing over all subjects of $\gX$.

\begin{figure}[tbh]
    \centering
    \includegraphics[width=\columnwidth]{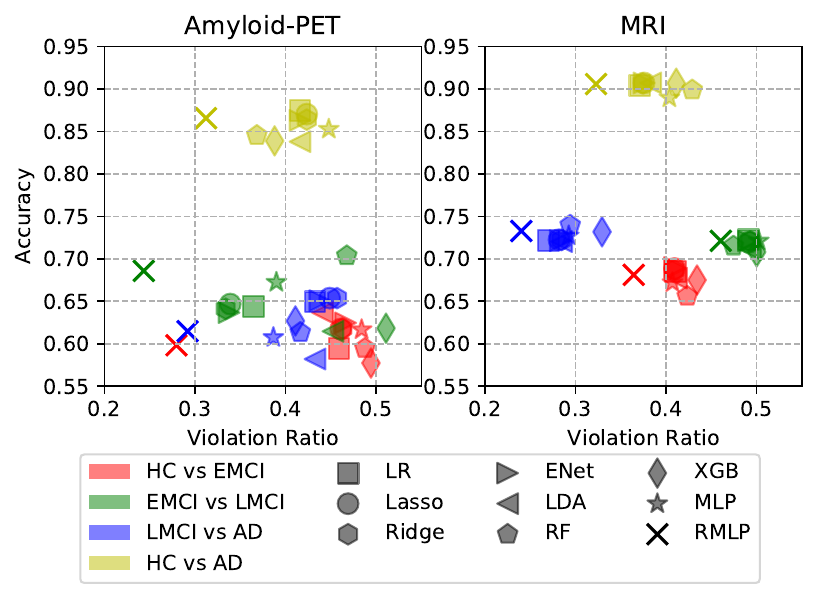}
    \vspace{-15pt}
    \caption{
    Illustration of the violation ratio -- accuracy of comparing methods on the Amyloid-PET (left) and the MRI (Right) datasets. The top left corner (higher accuracy, lower violation ratio) indicates better results.}
    \label{fig:Accuracy-VioR}
    \vspace{-5pt}
\end{figure}

\begin{figure*}
    \centering
    \includegraphics[width=0.9\textwidth]{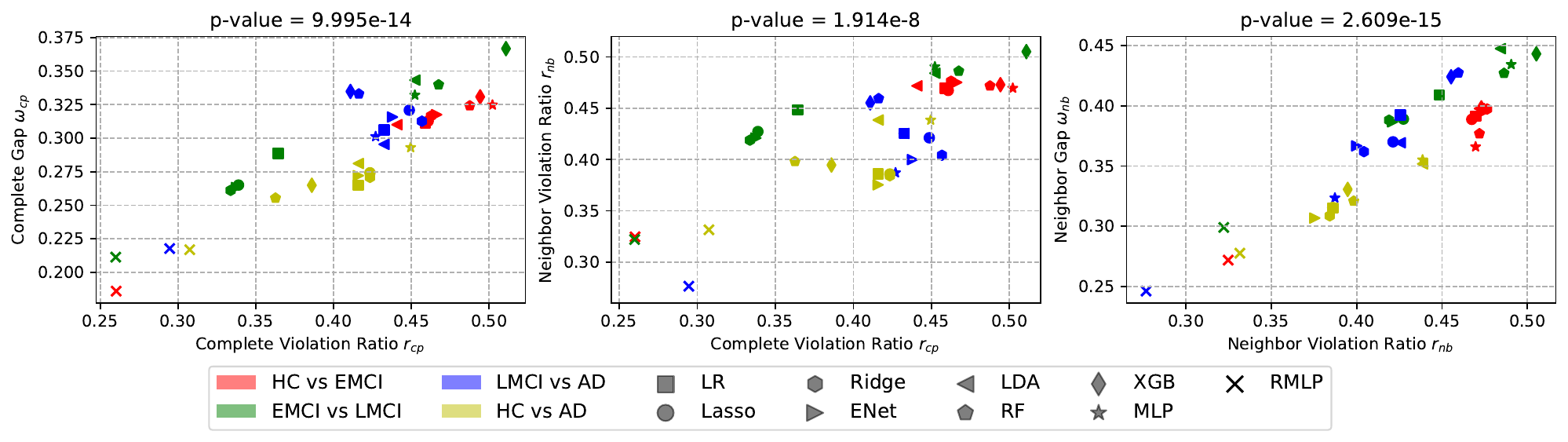}
    \caption{Illustration of the relation among neighbor/complete violation ratio/gap. They are represented by $r_{nb,cp},\omega_{nb,cp}$, respectively. Results show that the linear relation between the violation gap and the violation ratio is of great significance ($p=$9.995e-14, 2.609e-15). And the measurements based on the neighbor pairs are linearly consistent with complete pairs.
    }
    \vspace{-10pt}
    \label{fig:ratio-gap}
\end{figure*}

\vspace{-10pt}
\section{Experiments} 
\label{sec:experiments}
\vspace{-5pt}

In this section, we carry out experiments to show the effectiveness of the proposed regularization. First, for each dataset, 20\% of the data are held out randomly as the testing set, and 5-fold cross-validation experiments over the remained data are carried out. The datasets are split based on subjects $X_i$, which means different samples of the same subject do not appear in different subsets or even different folds. The baseline models being compared include linear regression (LR), Lasso, Ridge, elastic net (ENet), linear discriminative analysis (LDA), Random Forest (RF), XGBoost, and multilayer perceptron (MLP). Our proposed model is termed regularized MLP (RMLP). It shares the exact same structure as the MLP model tested. There are 6 hidden layers with $[64,64,64,32,32,32]$ neurons, respectively.

\vspace{-10pt}
\subsection{Expressiveness vs Violation Ratio}
\vspace{-5pt}

We first compare the expressiveness and the violation ratio of each model. The detailed accuracy and the violation ratio of both amyloid-PET and MRI data for all tested models are reported as tables in the appendix. Instead, here we visualize the results as accuracy--violation ratio in \cref{fig:Accuracy-VioR}, where different colors illustrate the four tasks, different shapes of markers represent different models, and the crossings represent the proposed RMLP. It is expected that the models should achieve both high accuracy and a low violation ratio. As a result, on the figure, they should be located the more closely to the top left corner, the better. For accuracy, the results show that due to the complexity of the tabular data, no model shows salient advantages over others and vice versa. However, when it comes to the violation ratio, RMLP outperforms others significantly. As demonstrated by their locations, overall the regularization achieves better monotonicity, while preserving comparable expressiveness.

\vspace{-5pt}
\subsection{Total Normalized Violation Gap}
\vspace{-5pt}

Note that the violation ratio is computed by counting the number of violation pairs, and dividing by the total number of pairs. This measurement might overlook local details -- how bad are the violations? When $f(\vx_{i,k+1})<f(\vx_{i,k})$, the larger the gap is, the more detrimental it is to the trustworthiness of the model.
Therefore, we compute the normalized violation gap for neighboring pairs
\begin{align}
    \begin{split}
    \omega_{nb} = \E_{X\sim\gX} \big[\sum_{k=1}^{|X|-1}&\frac{f(X_k) - f(X_{k+1})}{\max_{j<|X|}\big(f(X_j) - f(X_{j+1})\big)}\\
    &\bm{1}_{[f(X_{k+1})<f(X_k)]}\big]
    \end{split}
\end{align}
and also that for complete pairs $\omega_{cp}$. This is the sum of the violation gap between consecutive/all pairs of follow-up visits for all subjects, normalized by the maximum individual gap. The normalization is performed due to varying scales of predictions from different models. Besides, it should be noticed that it is the neighbor pairs that really determine the monotonicity. Therefore, we report the relation among neighbor/complete violation ratio $r_{nb},r_{cp}$ and neighbor/complete gap $\omega_{nb},\omega_{cp}$ of Amyloid-PET data in \cref{fig:ratio-gap}, following the same legend as before. It can be found that they show a strong linear correlation with $p$-value smaller than $10^{-7}$. This suggests that the discrete violation ratio is locally consistent with the violation gap and vice versa. 
{The results of MRI data can be found in the appendix.}

\begin{figure}[t]
    \centering
    \includegraphics[width=\columnwidth]{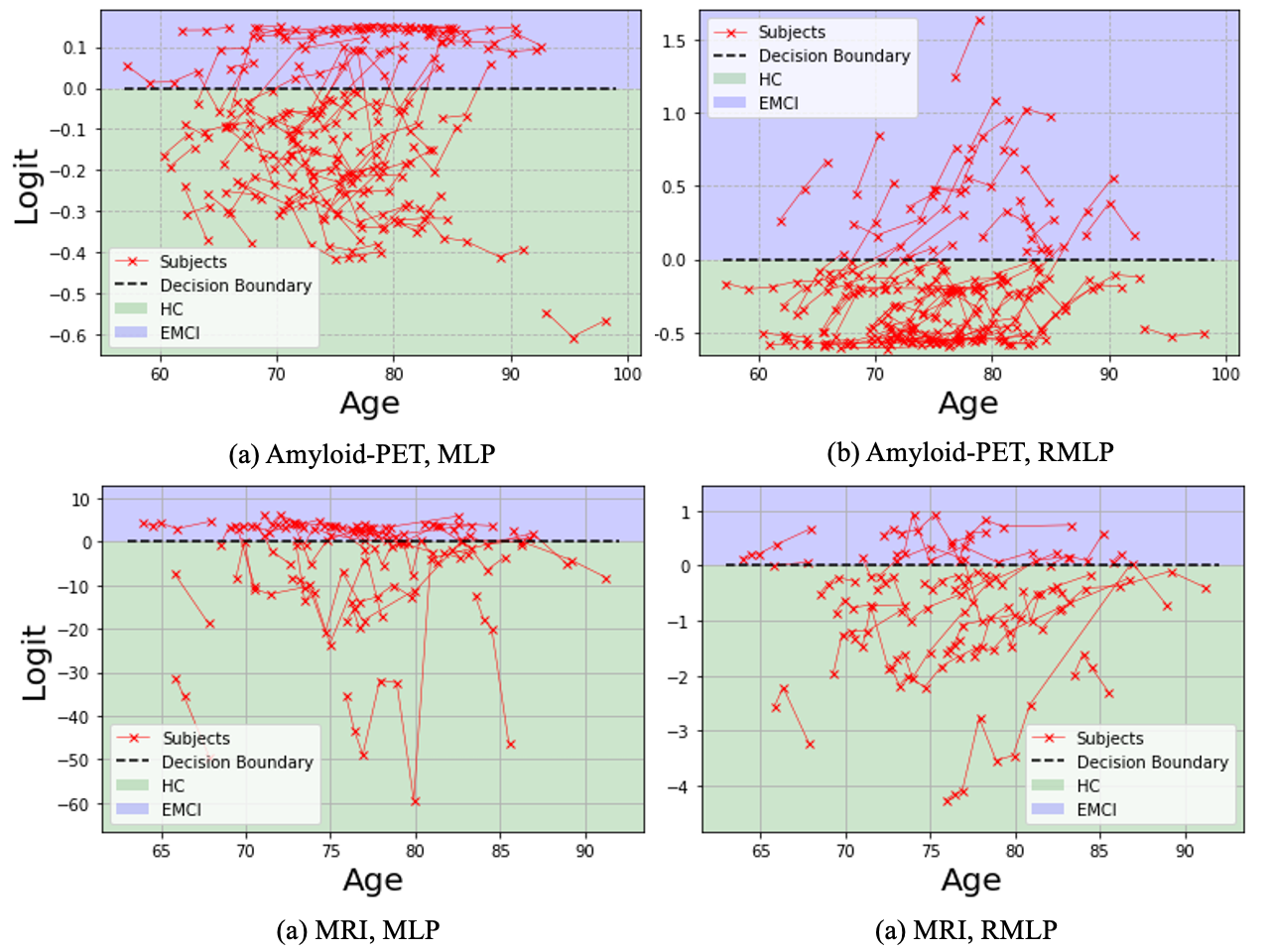}
    \vspace{-10pt}
    \caption{The progression trajectory of individual subjects learned from MLP (left) and RMLP (right) using Amyloid-PET (top) and MRI (bottom) test data. Red connected dots represent subjects with multiple follow-up visits. The black dashed horizontal line represents the decision boundary.}
    \vspace{-10pt}
    \label{fig:subject_trajectory}
\end{figure}

\vspace{-5pt}
\subsection{Visualizations of Trajectories}
\vspace{-5pt}

In \cref{fig:subject_trajectory}, we qualitatively compare the trajectories of test subjects for the task HC vs EMCI embedded by MLP (left) and RMLP(right), since they share the same structure and thereby illustrate the effectiveness of the regularization more clearly. Here we demonstrate the results from both amyloid-PET (top) and MRI (bottom) data. It can be found that although with similar performance on the accuracy, predicted risk by unregularized MLP is much more volatile across follow-up visits compared with RMLP. In particular, for subjects with multiple visits with the same diagnosis like HC or EMCI, the monotonicity of increasing risk is also preserved, suggesting the great potential of the proposed technique in modeling the subtle progression in the very early stage. The trajectories of other tasks follow the same trend, where the violation ratio is reduced similarly as shown in \cref{fig:Accuracy-VioR,fig:ratio-gap}.

\vspace{0pt}
\section{Conclusions}
\vspace{-5pt}

In this paper, we proposed a new regularization approach to help bridge a knowledge gap in the current research of AD. Specifically, when longitudinal data are taken as input independently, trained models tend to make predictions that violate the irreversibility of the AD progression across multiple visits of each subject. This undesired behavior undermines the trustworthiness of models in clinical applications even with high accuracy, and also limit further understanding of disease progression. To address this issue, we proposed a regularization approach that enforces the subject trajectories to align with the expected continuity in the increment of AD risk. We demonstrate the effectiveness of our approach through experiments on pre-analyzed Amyloid-PET and structural MRI tabular data, showing that our regularization improves the monotonicity of subject trajectories without sacrificing accuracy. This alignment with domain knowledge improves the consistency of the model's mechanism. Importantly our regularization approach is not specific to MLPs or to AD and can be applied to any task involving longitudinal data with prior knowledge of expected trends.

\clearpage
\bibliographystyle{IEEEbib}
{\footnotesize
\bibliography{main}

\begin{thebibliography}{10}

\bibitem{hampel2021amyloid}
Harald Hampel, John Hardy, Kaj Blennow, Christopher Chen, George Perry, Seung~Hyun Kim, Victor~L Villemagne, Paul Aisen, Michele Vendruscolo, Takeshi Iwatsubo, et~al.,
\newblock ``The amyloid-$\beta$ pathway in alzheimer’s disease,''
\newblock {\em Molecular psychiatry}, vol. 26, no. 10, pp. 5481--5503, 2021.

\bibitem{tong2017multi}
Tong Tong, Katherine Gray, Qinquan Gao, Liang Chen, Daniel Rueckert, Alzheimer's Disease~Neuroimaging Initiative, et~al.,
\newblock ``Multi-modal classification of alzheimer's disease using nonlinear graph fusion,''
\newblock {\em Pattern recognition}, vol. 63, pp. 171--181, 2017.

\bibitem{liu2012ensemble}
Manhua Liu, Daoqiang Zhang, Dinggang Shen, Alzheimer's Disease~Neuroimaging Initiative, et~al.,
\newblock ``Ensemble sparse classification of alzheimer's disease,''
\newblock {\em NeuroImage}, vol. 60, no. 2, pp. 1106--1116, 2012.

\bibitem{wisnu2022xadlime}
Ahmad Wisnu~Mulyadi, Wonsik Jung, Kwanseok Oh, Jee~Seok Yoon, and Heung-Il Suk,
\newblock ``Xadlime: explainable alzheimer's disease likelihood map estimation via clinically-guided prototype learning,''
\newblock {\em arXiv e-prints}, pp. arXiv--2207, 2022.

\bibitem{wang2022deep}
Qi~Wang, Kewei Chen, Yi~Su, Eric~M Reiman, Joel~T Dudley, and Benjamin Readhead,
\newblock ``Deep learning-based brain transcriptomic signatures associated with the neuropathological and clinical severity of alzheimer’s disease,''
\newblock {\em Brain Communications}, vol. 4, no. 1, pp. fcab293, 2022.

\bibitem{basaia2019automated}
Silvia Basaia, Federica Agosta, Luca Wagner, Elisa Canu, Giuseppe Magnani, Roberto Santangelo, Massimo Filippi, Alzheimer's Disease~Neuroimaging Initiative, et~al.,
\newblock ``Automated classification of alzheimer's disease and mild cognitive impairment using a single mri and deep neural networks,''
\newblock {\em NeuroImage: Clinical}, vol. 21, pp. 101645, 2019.

\bibitem{ouyang2020longitudinal}
Jiahong Ouyang, Qingyu Zhao, Edith~V Sullivan, Adolf Pfefferbaum, Susan~F Tapert, Ehsan Adeli, and Kilian~M Pohl,
\newblock ``Longitudinal pooling \& consistency regularization to model disease progression from mris,''
\newblock {\em IEEE journal of biomedical and health informatics}, vol. 25, no. 6, pp. 2082--2092, 2020.

\bibitem{tasaki2022inferring}
Shinya Tasaki, Jishu Xu, Denis~R Avey, Lynnaun Johnson, Vladislav~A Petyuk, Robert~J Dawe, David~A Bennett, Yanling Wang, and Chris Gaiteri,
\newblock ``Inferring protein expression changes from mrna in alzheimer’s dementia using deep neural networks,''
\newblock {\em Nature Communications}, vol. 13, no. 1, pp. 655, 2022.

\bibitem{ross2017right}
Andrew~Slavin Ross, Michael~C Hughes, and Finale Doshi-Velez,
\newblock ``Right for the right reasons: Training differentiable models by constraining their explanations,''
\newblock {\em arXiv preprint arXiv:1703.03717}, 2017.

\bibitem{rieger2020interpretations}
Laura Rieger, Chandan Singh, William Murdoch, and Bin Yu,
\newblock ``Interpretations are useful: penalizing explanations to align neural networks with prior knowledge,''
\newblock in {\em International conference on machine learning}. PMLR, 2020, pp. 8116--8126.

\bibitem{boopathy2020proper}
Akhilan Boopathy, Sijia Liu, Gaoyuan Zhang, Cynthia Liu, Pin-Yu Chen, Shiyu Chang, and Luca Daniel,
\newblock ``Proper network interpretability helps adversarial robustness in classification,''
\newblock in {\em International Conference on Machine Learning}. PMLR, 2020, pp. 1014--1023.

\bibitem{wang2021self}
Yipei Wang and Xiaoqian Wang,
\newblock ``Self-interpretable model with transformation equivariant interpretation,''
\newblock {\em Advances in Neural Information Processing Systems}, vol. 34, pp. 2359--2372, 2021.

\bibitem{jack2010update}
Clifford~R Jack~Jr, Matt~A Bernstein, Bret~J Borowski, Jeffrey~L Gunter, Nick~C Fox, Paul~M Thompson, Norbert Schuff, Gunnar Krueger, Ronald~J Killiany, Charles~S DeCarli, et~al.,
\newblock ``Update on the magnetic resonance imaging core of the alzheimer's disease neuroimaging initiative,''
\newblock {\em Alzheimer's \& Dementia}, vol. 6, no. 3, pp. 212--220, 2010.

\bibitem{saykin2010alzheimer}
Andrew~J Saykin, Li~Shen, Tatiana~M Foroud, Steven~G Potkin, Shanker Swaminathan, Sungeun Kim, Shannon~L Risacher, Kwangsik Nho, Matthew~J Huentelman, David~W Craig, et~al.,
\newblock ``Alzheimer's disease neuroimaging initiative biomarkers as quantitative phenotypes: Genetics core aims, progress, and plans,''
\newblock {\em Alzheimer's \& Dementia}, vol. 6, no. 3, pp. 265--273, 2010.

\bibitem{weiner2013alzheimer}
Michael~W Weiner, Dallas~P Veitch, Paul~S Aisen, Laurel~A Beckett, Nigel~J Cairns, Robert~C Green, Danielle Harvey, Clifford~R Jack, William Jagust, Enchi Liu, et~al.,
\newblock ``The alzheimer's disease neuroimaging initiative: a review of papers published since its inception,''
\newblock {\em Alzheimer's \& Dementia}, vol. 9, no. 5, pp. e111--e194, 2013.

\bibitem{edmonds2016patterns}
Emily~C Edmonds, Katherine~J Bangen, Lisa Delano-Wood, Daniel~A Nation, Ansgar~J Furst, David~P Salmon, Mark~W Bondi, Alzheimer’s Disease~Neuroimaging Initiative, et~al.,
\newblock ``Patterns of cortical and subcortical amyloid burden across stages of preclinical alzheimer’s disease,''
\newblock {\em Journal of the International Neuropsychological Society}, vol. 22, no. 10, pp. 978--990, 2016.

\bibitem{oh2019classification}
Kanghan Oh, Young-Chul Chung, Ko~Woon Kim, Woo-Sung Kim, and Il-Seok Oh,
\newblock ``Classification and visualization of alzheimer’s disease using volumetric convolutional neural network and transfer learning,''
\newblock {\em Scientific Reports}, vol. 9, no. 1, pp. 1--16, 2019.

\bibitem{zhou2019effective}
Tao Zhou, Kim-Han Thung, Xiaofeng Zhu, and Dinggang Shen,
\newblock ``Effective feature learning and fusion of multimodality data using stage-wise deep neural network for dementia diagnosis,''
\newblock {\em Human brain mapping}, vol. 40, no. 3, pp. 1001--1016, 2019.

\bibitem{you2020alzheimer}
Zeng You, Runhao Zeng, Xiaoyong Lan, Huixia Ren, Zhiyang You, Xue Shi, Shipeng Zhao, Yi~Guo, Xin Jiang, and Xiping Hu,
\newblock ``Alzheimer's disease classification with a cascade neural network,''
\newblock {\em Frontiers in Public Health}, vol. 8, pp. 584387, 2020.

\bibitem{odusami2021analysis}
Modupe Odusami, Rytis Maskeli{\=u}nas, Robertas Dama{\v{s}}evi{\v{c}}ius, and Tomas Krilavi{\v{c}}ius,
\newblock ``Analysis of features of alzheimer’s disease: detection of early stage from functional brain changes in magnetic resonance images using a finetuned resnet18 network,''
\newblock {\em Diagnostics}, vol. 11, no. 6, pp. 1071, 2021.

\bibitem{shwartz2022tabular}
Ravid Shwartz-Ziv and Amitai Armon,
\newblock ``Tabular data: Deep learning is not all you need,''
\newblock {\em Information Fusion}, vol. 81, pp. 84--90, 2022.

\bibitem{chen2016xgboost}
Tianqi Chen and Carlos Guestrin,
\newblock ``Xgboost: A scalable tree boosting system,''
\newblock in {\em Proceedings of the 22nd acm sigkdd international conference on knowledge discovery and data mining}, 2016, pp. 785--794.

\bibitem{rumelhart1985learning}
David~E Rumelhart, Geoffrey~E Hinton, and Ronald~J Williams,
\newblock ``Learning internal representations by error propagation,''
\newblock Tech. {R}ep., California Univ San Diego La Jolla Inst for Cognitive Science, 1985.

\bibitem{gorishniy2021revisiting}
Yury Gorishniy, Ivan Rubachev, Valentin Khrulkov, and Artem Babenko,
\newblock ``Revisiting deep learning models for tabular data,''
\newblock {\em Advances in Neural Information Processing Systems}, vol. 34, pp. 18932--18943, 2021.

\bibitem{ebrahimighahnavieh2020deep}
Mr~Amir Ebrahimighahnavieh, Suhuai Luo, and Raymond Chiong,
\newblock ``Deep learning to detect alzheimer's disease from neuroimaging: A systematic literature review,''
\newblock {\em Computer methods and programs in biomedicine}, vol. 187, pp. 105242, 2020.

\bibitem{wen2020convolutional}
Junhao Wen, Elina Thibeau-Sutre, Mauricio Diaz-Melo, Jorge Samper-Gonz{\'a}lez, Alexandre Routier, Simona Bottani, Didier Dormont, Stanley Durrleman, Ninon Burgos, Olivier Colliot, et~al.,
\newblock ``Convolutional neural networks for classification of alzheimer's disease: Overview and reproducible evaluation,''
\newblock {\em Medical image analysis}, vol. 63, pp. 101694, 2020.

\bibitem{saleem2022deep}
Tausifa~Jan Saleem, Syed~Rameem Zahra, Fan Wu, Ahmed Alwakeel, Mohammed Alwakeel, Fathe Jeribi, and Mohammad Hijji,
\newblock ``Deep learning-based diagnosis of alzheimer’s disease,''
\newblock {\em Journal of Personalized Medicine}, vol. 12, no. 5, pp. 815, 2022.

\end{thebibliography}
}

\end{document}